\documentclass[10pt,twocolumn,letterpaper]{article}

\usepackage{cvpr}              

\definecolor{cvprblue}{rgb}{0.21,0.49,0.74}
\usepackage[pagebackref,breaklinks,colorlinks,allcolors=cvprblue]{hyperref}
\usepackage[utf8]{inputenc} 
\usepackage[T1]{fontenc}    
\usepackage{hyperref}       
\usepackage{url}            
\usepackage{booktabs}       
\usepackage{amsfonts}       
\usepackage{nicefrac}       
\usepackage{microtype}      
\usepackage{xcolor}         
\usepackage{multirow}
\usepackage{array}
\usepackage{amsmath}
\usepackage{diagbox}
\usepackage{colortbl}
\usepackage{graphicx}
\usepackage{float}
\usepackage{subcaption}
\usepackage{tabularx}
\usepackage{soul}
\usepackage{placeins}
\usepackage{algorithm}
\usepackage{algpseudocode}

\def\paperID{9364} 
\def\confName{CVPR}
\def\confYear{2025}

\def\romannum#1{\uppercase\expandafter{\romannumeral#1}}

\title{SSHNet: Unsupervised Cross-modal Homography Estimation \\ via Problem Reformulation and Split Optimization}

\author{
	Junchen Yu$^{1,2}$ \quad Si-Yuan Cao$^{1,2,3}$\thanks{Corresponding author.} \quad Runmin Zhang$^{2}$ \quad Chenghao Zhang$^{2}$ \\ 
	Zhu Yu$^{2}$ \quad Shujie Chen$^{4}$ \quad Bailin Yang$^{4}$ \quad Hui-Liang Shen$^{2}$ \\
	\small $^{1}$Ningbo Innovation Center, Zhejiang University \quad $^{2}$College of Information Science and Electronic Engineering, Zhejiang University \\
	\small $^{3}$ NingboTech University \quad $^{4}$ Zhejiang Key Laboratory of Big Data and Future ECommerce Technology, Hangzhou, China \\
	{\tt\small \{yujunchen,cao\_siyuan,runmin\_zhang,zch00,yu\_zhu\}@zju.edu.cn} \\ {\tt\small \{chenshujie,ybl\}@zjgsu.edu.cn \quad shenhl@zju.edu.cn}
}

\begin{document}
\maketitle
\begin{abstract}
We propose a novel unsupervised cross-modal homography estimation learning framework, named \textbf{S}plit \textbf{S}upervised \textbf{H}omography estimation \textbf{N}etwork (SSHNet). SSHNet reformulates the unsupervised cross-modal homography estimation into two supervised sub-problems, each addressed by its specialized network: a homography estimation network and a modality transfer network. To realize stable training, we introduce an effective split optimization strategy to train each network separately within its respective sub-problem. We also formulate an extra homography feature space supervision to enhance feature consistency, further boosting the estimation accuracy. Moreover, we employ a simple yet effective distillation training technique to reduce model parameters and improve cross-domain generalization ability while maintaining comparable performance. 
The training stability of SSHNet enables its cooperation with various homography estimation architectures. Experiments reveal that the SSHNet using IHN as homography estimation network, namely SSHNet-IHN, outperforms previous unsupervised approaches by a significant margin. Even compared to supervised approaches MHN and LocalTrans, SSHNet-IHN achieves 47.4\% and 85.8\% mean average corner errors (MACEs) reduction on the challenging OPT-SAR dataset. Source code is available at \url{https://github.com/Junchen-Yu/SSHNet}.
\end{abstract}    
\section{Introduction}
\label{sec:intro}

Cross-modal homography estimation aims to solve the global perspective transformation between two images under different modalities. It is widely used in various computer vision tasks, such as GPS-denied robotic localization~\cite{goforth2019gps, wang2023fine}, multi-modal image restoration~\cite{marivani2022designing, dharejo2022multimodal}, and multi-spectral image fusion~\cite{zhou2019integrated, ying2021unaligned}. Recent supervised homography estimation approaches~\cite{detone2016deep, le2020deep, shao2021localtrans, zhao2021deep, cao2022iterative, cao2023recurrent} have demonstrated their superiority in handling large deformation and modality gaps. However, since multi-modal images are captured by different sensors, the ground-truth homography deformations are usually unknown in real-world applications.

\begin{figure}[t]
    \centering
    \subfloat{
        \includegraphics[width=0.082\textwidth]{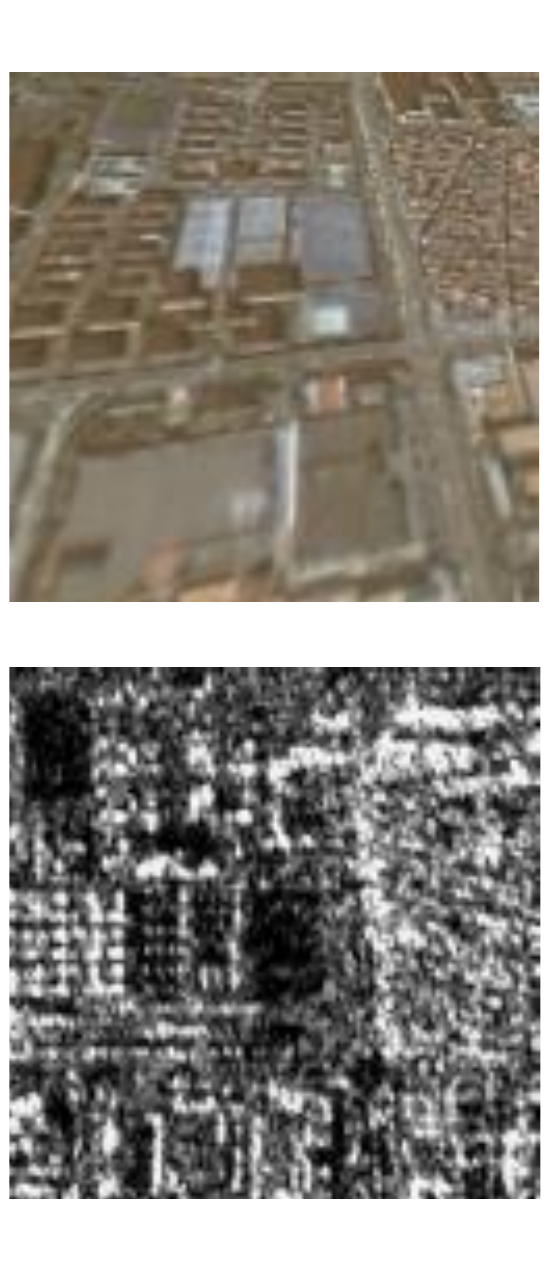}}
    \subfloat{
        \includegraphics[width=0.39\textwidth]{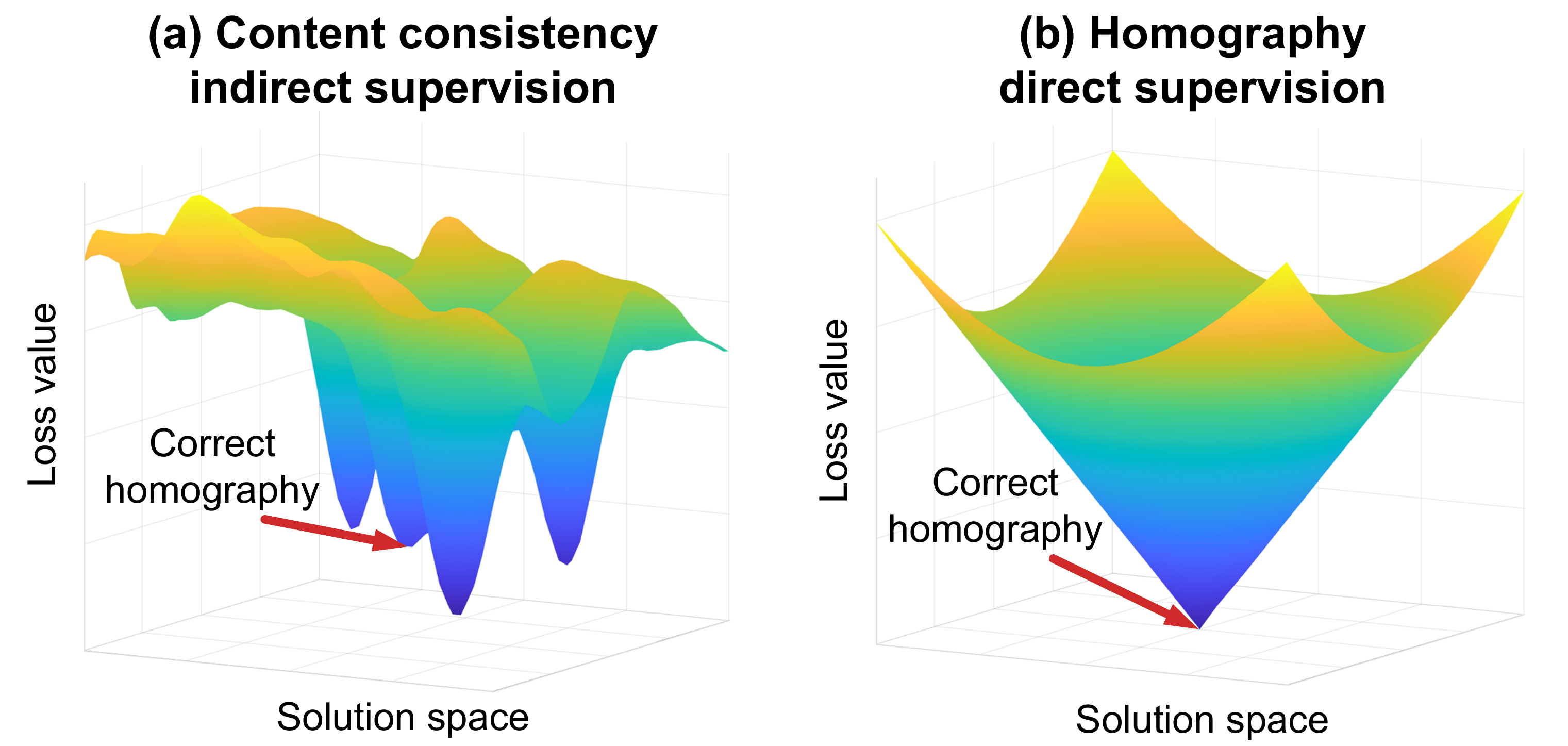}}
    \vspace{-0.7em}
    \caption{
        Comparison of the solution space between (a) Content consistency indirect supervision and (b) Homography direct supervision. The content consistency indirect supervision may mislead the training due to local minima, while homography direct supervision won't.
    }\label{fig:intro}
    \vspace{-1.5em}
\end{figure}

To avoid using labeled data, unsupervised homography estimation approaches have emerged. Existing approaches~\cite{nguyen2018unsupervised,zhang2020content, ye2021motion, koguciuk2021perceptual} typically optimize the similarity between the warped source image and target image through intensity-based losses. Some multi-modal image registration approaches~\cite{arar2020unsupervised, wang2022unsupervised} transferred an image into the other modality, using mono-modal similarity metrics on image content to measure the correctness of motion. The previous state-of-the-art (SOTA) approach~\cite{zhang2024scpnet} combined cross-modal intensity-based learning with intra-modal self-supervised learning for homography knowledge injection. Notably, almost all the previous approaches fully or partially rely on the content consistency of images or feature maps to indirectly assess the correctness of homography estimation. As illustrated in Fig.~\ref{fig:intro}a, due to the non-convexity of the image signal, loss value of content consistency indirect supervision has multiple local minima across its solution space, and hence can not guarantee accurate homography prediction, especially under large modality and deformation gaps. On the contrary, the homography direct supervision in Fig.~\ref{fig:intro}b can avoid such drawback, but it is unavailable in unsupervised cross-modal homography estimation, making the problem more challenging.

To cope with the above problem, we propose a novel unsupervised cross-modal homography estimation learning framework, named \textbf{S}plit \textbf{S}upervised \textbf{H}omography estimation \textbf{N}etwork (SSHNet). SSHNet reformulates the unsupervised cross-modal homography estimation problem into two imperfect sub-problems: \textbf{Sub-problem \romannum{1}:} A supervised mono-modal homography estimation problem, where the modalities of input images are not perfectly consistent\@. \textbf{Sub-problem \romannum{2}:} A supervised modality transfer problem, where the supervisory image is not perfectly aligned. We introduce a split optimization strategy to separately train a homography estimation network for Sub-problem \romannum{1} and a modality transfer network for Sub-problem \romannum{2}, enabling effective convergence of the framework. Thanks to the proposed problem reformulation and the well-designed optimization strategy, our framework effectively solves the unsupervised task by training the networks under totally direct supervision, showing significantly improved training stability. This enables the successful training of iterative homography estimation architectures like IHN~\cite{cao2022iterative} and RHWF~\cite{cao2023recurrent}. Better still, for the correlation-based homography estimation architectures~\cite{cao2022iterative,zhang2024scpnet,cao2023recurrent}, we propose an extra homography feature space supervision that enhances feature consistency, further improving homography estimation accuracy. Moreover, to address the computational burden of the modality transfer network, we introduce a simple yet effective distillation training technique, which notably reduces model parameters and improves cross-domain generalization ability.

We evaluate SSHNet on a variety of datasets including GoogleMap~\cite{zhao2021deep}, DPDN~\cite{riegler2016deep}, OPT-SAR~\cite{li2022mcanet}, and Flash/no-flash~\cite{he2014saliency} cross-modal datasets, together with RGB/NIR~\cite{brown2011multi} cross-spectral datasets. SSHNet outperforms all the previous unsupervised homography estimation approaches with a large gap, and even achieves better performance compared to many supervised ones.
The SSHNet using IHN as homography estimation network, namely SSHNet-IHN, owns 47.4\% and 85.8\% lower mean average corner errors (MACEs) than the supervised approaches MHN~\cite{le2020deep} and LocalTrans~\cite{shao2021localtrans} on the challenging OPT-SAR dataset. Moreover, the SSHNet using the homography estimation architecture of the previous unsupervised SOTA approach SCPNet~\cite{zhang2024scpnet}, namely SSHNet-SCPNet, even outperforms SCPNet itself, revealing the effectiveness of our proposed learning framework. 
To summarize, our main contributions are as follows:

\begin{itemize}
    \item {
        We propose SSHNet, a novel unsupervised cross-modal homography estimation learning framework, which reformulates the unsupervised problem into two sub-problems with direct supervision. To realize stable training, we introduce a split optimization strategy that separately trains the network in each sub-problem.
    }
    \item {
        For correlation-based homography estimation architectures, we introduce an extra homography feature space supervision to improve feature consistency for homography estimation, which further improves estimation accuracy.
    }
    \item {
        Considering the computational burden introduced by the additional modality transfer network, we devise a simple yet effective distillation training technique to reduce model parameters 
        while improving cross-domain generalization ability with little impact on performance.
    }
    \item {
        The training stability of SSHNet enables its cooperation with various homography estimation architectures, especially iterative ones like IHN and RHWF, resulting in higher performance. SSHNet-IHN outperforms the previous unsupervised approaches by a significant margin.
    }
\end{itemize}
\section{Related Works}\label{sec:related}


\textbf{Supervised Deep Homography Estimation.} 
DeTone \emph{et al.}~\cite{detone2016deep} introduced DHN, a VGG-style network to predict the homography between the source and target images.
Le \emph{et al.}~\cite{le2020deep} introduced a multi-scale VGG-style network, named MHN. Chang \emph{et al.}~\cite{chang2017clkn} combined CNN with an untrainable IC-LK iterator. Zhao \emph{et al.}~\cite{zhao2021deep} enhanced the feature similarity to improve the performance of IC-LK iterator. Shao \emph{et al.}~\cite{shao2021localtrans} proposed LocalTrans that adopts a multi-scale local transformer to register cross-resolution images. Cao \emph{et al.}~\cite{cao2022iterative} proposed IHN, a trainable iterative framework with substantial accuracy. Cao \emph{et al.}~\cite{cao2023recurrent} further introduced RHWF that integrates homography-guided image warping and focus transformer, achieving higher performance. Deng \emph{et al.}~\cite{deng2024crosshomo} proposed CrossHomo, a multi-level framework for cross-modality and cross-resolution homography estimation. Li \emph{et al.}~\cite{li2024dmhomo} introduced DMHomo, a diffusion-based framework to generate high-quality training pairs for supervised homography learning.

\textbf{Unsupervised Deep Homography Estimation.} 
Nguyen \emph{et al.}~\cite{nguyen2018unsupervised} proposed UDHN, an unsupervised homography estimation network trained with photometric loss. Zhang \emph{et al.}~\cite{zhang2020content} introduced CA-UDHN, projecting images into a consistent feature space for effective estimation. Koguchiuk \emph{et al.}\cite{koguciuk2021perceptual} proposed biHomE, using perceptual loss\cite{johnson2016perceptual} to improve robustness to intensity and viewpoint variations. Ye \emph{et al.}~\cite{ye2021motion} proposed BasesHomo, which enforces the image feature to be warp-equivalent with a feature identity loss and decomposes the homography matrix into flow bases. As modality and deformation gaps grow, intensity-based metrics for cross-modal homography matching may fail~\cite{zhang2024scpnet}. To address this, Zhang \emph{et al.}~\cite{zhang2024scpnet} introduced SCPNet, which combines cross-modal intensity-based learning with intra-modal self-supervised learning for homography knowledge injection. Besides, some unsupervised approaches, such as NeMAR~\cite{arar2020unsupervised}, RFNet~\cite{xu2022rfnet}, and UMF-CMGR~\cite{wang2022unsupervised}, use modality transfer networks to convert one modality into the other, enabling cross-modal motion estimation. The above unsupervised approaches perform well on images with small deformations, but may produce unsatisfactory results under large deformations and modality gaps.

\textbf{Image-to-Image Translation.} 
The image-to-image (I2I) translation problem focuses on converting images from a source domain to a target domain.
Isola \emph{et al.}~\cite{isola2017image} introduced Pix2pix, which employs conditional GANs and requires paired training data. Zhu \emph{et al.}~\cite{zhu2017unpaired} later proposed CycleGAN, utilizing cycle-consistent loss to enable unpaired I2I translation. Liu \emph{et al.}\cite{liu2017unsupervised} developed UNIT, and Huang \emph{et al.}\cite{huang2018multimodal} introduced MUNIT, both leveraging a shared latent space assumption to improve unpaired translation. More recently, Sasaki \emph{et al.}\cite{sasaki2021unit} proposed UNIT-DDPM, integrating diffusion probabilistic models into the I2I translation framework. Li \emph{et al.}~\cite{li2023bbdm} introduced BBDM, modeling I2I translation as a stochastic Brownian Bridge process.
\begin{figure}[t]
	\centering
	\includegraphics[width=0.47\textwidth]{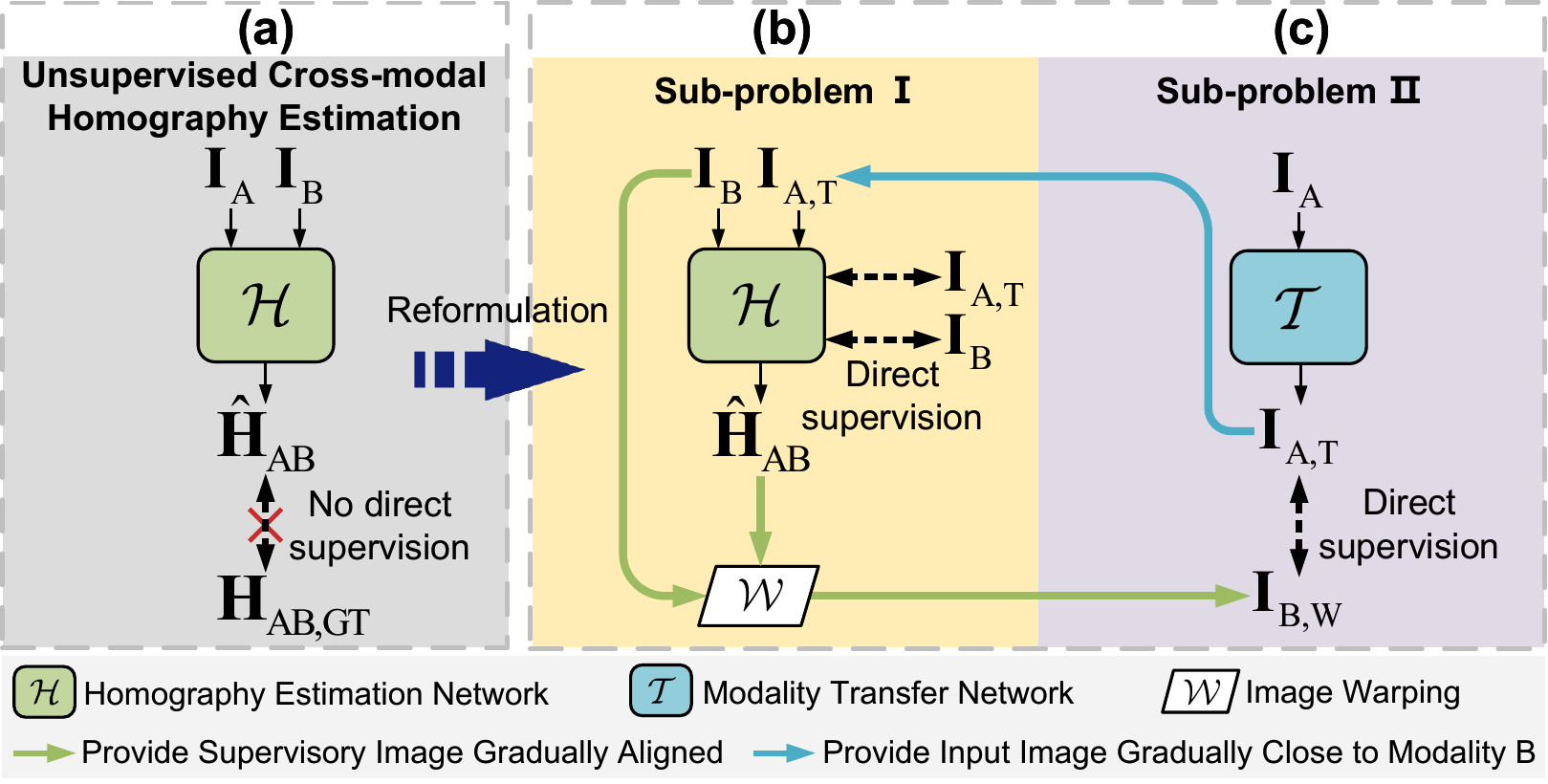}
	\vspace{-0.5em}
	\caption{Reformulating the unsupervised cross-modal homography estimation problem into two sub-problems with direct supervision. (a) Unsupervised cross-modal homography estimation problem without direct supervision. (b) Mono-modal homography estimation problem with imperfect inputs. (c) Modality transfer problem with imperfect supervision.}\label{fig:prob_redef}
	\vspace{-1.5em}
\end{figure}

\section{Problem Reformulation}\label{sec:reformulation} 
Denote $\mathbf{I}_{\mathrm{A}}$ and $\mathbf{I}_{\mathrm{B}}$ as a pair of images from modality A and B, with an unknown homography $\mathbf{H}_{\mathrm{AB,GT}}$. As illustrated in Fig.~\ref{fig:prob_redef}a, the unsupervised cross-modal homography estimation aims to train a network to estimate the homography $\hat{\mathbf{H}}_{\mathrm{AB}}$, without direct supervision of $\mathbf{H}_{\mathrm{AB,GT}}$. To solve this problem, in SSHNet, we reformulate it into two sub-problems with direct supervision, including a mono-modal homography estimation problem with imperfect inputs (Fig.~\ref{fig:prob_redef}b) and a modality transfer problem with imperfect supervision (Fig.~\ref{fig:prob_redef}c). The two sub-problems are mutually causal, and the imperfect condition of each sub-problem can be gradually addressed through the optimization of another sub-problem. In this way, the unsupervised homography estimation problem can be solved in a totally supervised manner.

\textbf{Sub-problem \romannum{1}.} For the mono-modal homography estimation, generating image pairs with simulated ground-truth deformations for the direct self-supervised training is a common approach~\cite{detone2016deep,le2020deep,cao2022iterative,shao2021localtrans}.
If we can ideally convert $\mathbf{I}_\mathrm{A}$ into modality $\mathrm{B}$, producing the transferred image $\mathbf{I}_{\mathrm{A,T}}$, we can treat the unsupervised cross-modal homography estimation problem as the mono-modal one by estimating the homography between $\mathbf{I}_{\mathrm{A,T}}$ and $\mathbf{I}_{\mathrm{B}}$, as in Fig.~\ref{fig:prob_redef}b. The homography estimation network can be effectively trained using $\mathbf{I}_{\mathrm{B}}$ with the simulated homography deformations, in a self-supervised manner. We refer to the problem of estimating the homography between $\mathbf{I}_{\mathrm{B}}$ and the transferred image $\mathbf{I}_{\mathrm{A,T}}$ with the self-supervised training using $\mathbf{I}_{\mathrm{B}}$ as Sub-problem \romannum{1}. 

\textbf{Sub-problem \romannum{2}.} As illustrated in Fig.~\ref{fig:prob_redef}c, we also need a modality transfer network to obtain the optimal mapping from $\mathbf{I}_{\mathrm{A}}$ to $\mathbf{I}_{\mathrm{B}}$, producing $\mathbf{I}_{\mathrm{A,T}}$. The ideal network training typically requires $\mathbf{I}_{\mathrm{B}}$ that is well-aligned with $\mathbf{I}_{\mathrm{A}}$ for supervision, which is also unavailable in the unsupervised cross-modal homography estimation. 
Previous approaches~\cite{wang2022unsupervised, arar2020unsupervised} attempt to use GAN-based supervision to achieve modality transfer training using the unaligned image pairs. However, this strategy encounters considerable challenges when dealing with large deformation and modality gaps, as illustrated in~\cite{zhang2024scpnet}. Therefore, high-quality modality transfer requires well-aligned $\mathbf{I}_{\mathrm{A}}$ and $\mathbf{I}_{\mathrm{B}}$. Now that we have transferred $\mathbf{I}_{\mathrm{A}}$ to $\mathbf{I}_{\mathrm{A,T}}$, the requirement becomes that Sub-problem \romannum{1} being ideally solved. 
We refer to the modality transfer problem, which uses the homography estimated in Sub-problem \romannum{1} to produce a warped $\mathbf{I}_{\mathrm{B}}$ for supervision, as Sub-problem \romannum{2}.

\begin{figure}[t]
    \centering
	\includegraphics[width=0.48\textwidth]{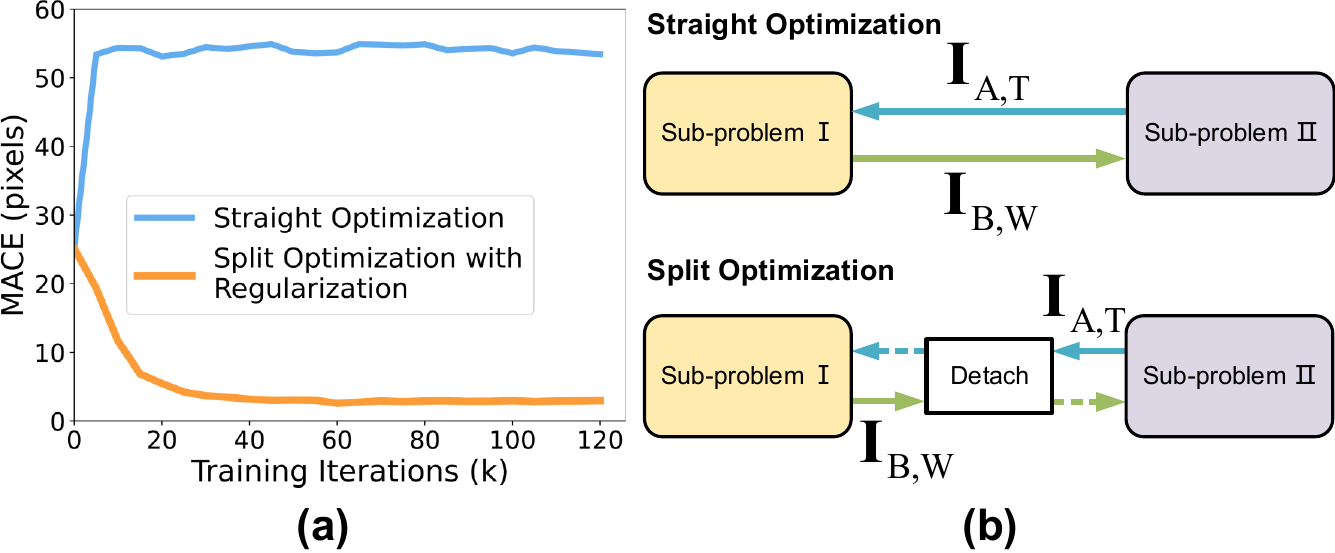}
    \vspace{-1.5em}
    \caption{
        Comparison of straight optimization and split optimization with regularization. (a) Cross-modal test MACEs for the network using straight optimization and split optimization with regularization, on OPT-SAR dataset. (b) Schematic diagram of straight optimization and split optimization.
    }\label{fig:opt_MACE}
    \vspace{-1.5em}
\end{figure}

\textbf{Straight Optimization Fails.} However, straight optimization of both sub-problems simultaneously is challenging, as the settlement of either sub-problem depends on the assumption that the other has already been solved. We conduct a pilot experiment where the homography estimation network and the modality transfer network are trained simultaneously to address Sub-problem \romannum{1} and \romannum{2}, but the networks fail to converge, as shown by the blue curve in Fig.~\ref{fig:opt_MACE}a.

\begin{figure*}[t]
	\centering
	\includegraphics[width=\textwidth]{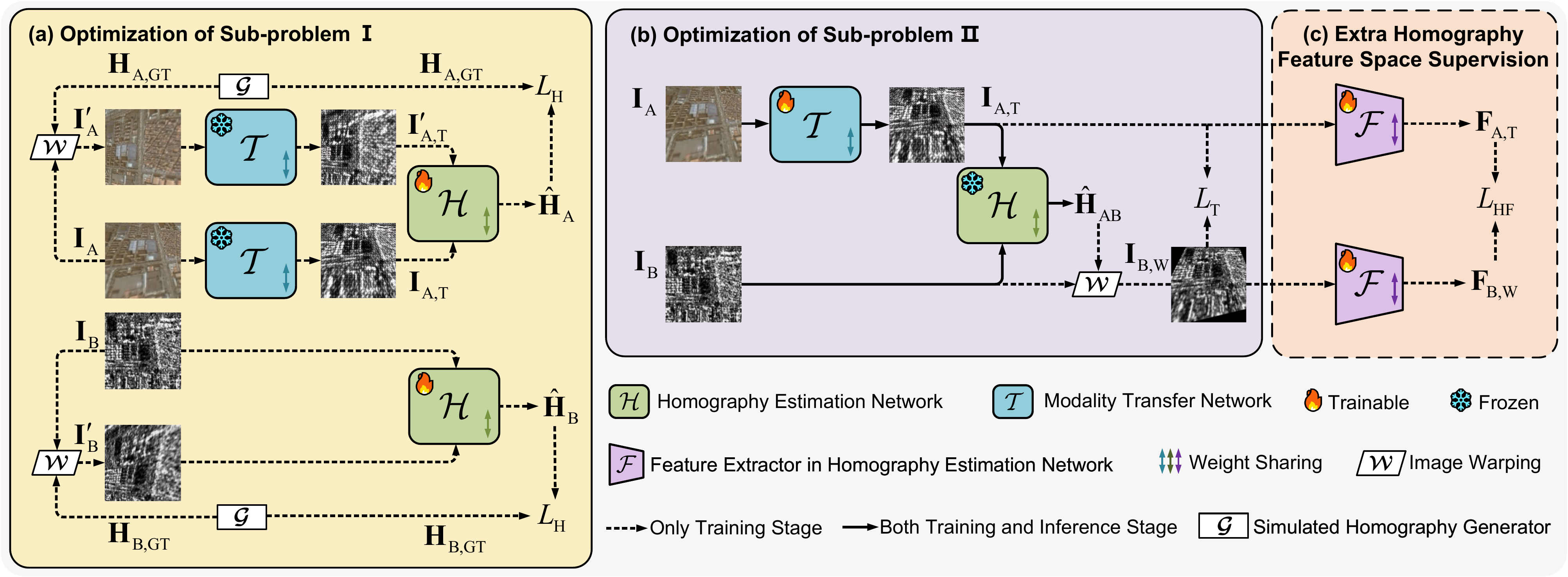}
	\caption{
		Schematic diagram of \textbf{S}plit \textbf{S}upervised \textbf{H}omography estimation \textbf{N}etwork, namely SSHNet\@. (a) Detailed optimization process of Sub-problem \romannum{1}\@. (b) Detailed optimization process of Sub-problem \romannum{2}\@. (c) Extra supervision on the homography feature space.
	}\label{fig:method}
	\vspace{-1.5em}
\end{figure*}

\textbf{Split Optimization with Regularization Success.} Inspired by decomposition methods in convex optimization, we then consider using a split optimization strategy to address the two sub-problems separately. 
During optimization, each network updates its weights with the other one's frozen. As in Fig.~\ref{fig:prob_redef}b and Fig.~\ref{fig:prob_redef}c, in each iteration, the homography estimation network in Sub-problem \romannum{1} is updated, producing a better-aligned $\mathbf{I}_{\mathrm{B}}$ to guide the modality transfer network in Sub-problem \romannum{2} in the next iteration. Similarly, the modality transfer network in Sub-problem \romannum{2} is updated, generating a better transferred image $\mathbf{I}_{\mathrm{A,T}}$ to serve as input for the homography estimation network in Sub-problem \romannum{1}. 
This process gradually improves the outputs of each network during training, facilitating the optimization of the other network, and finally makes both sub-problems nearly perfect supervised ones. 
However, at the beginning of the training, the modality gap between $\mathbf{I}_{\mathrm{A,T}}$ and $\mathbf{I}_{\mathrm{B}}$ in Sub-problem \romannum{1} is extremely large, so is the misalignment of $\mathbf{I}_{\mathrm{A}}$ and $\mathbf{I}_{\mathrm{B}}$ in Sub-problem \romannum{2}, which hinders the mutual promotion between two sub-problems. A regularization term that guides the training into our desired direction is needed. Inspired by~\cite{zhang2024scpnet}~\footnote{\cite{zhang2024scpnet} finds that if training the weight-shared homography estimation network using the simulated deformation within each two modalities, the cross-modal homography knowledge can be implicitly learned through the multitask learning mechanism~\cite{caruana1997multitask, doersch2017multi}.}, in Sub-problem \romannum{1}, we replace the self-supervised training using only $\mathbf{I}_{\mathrm{B}}$ by the two-branch self-supervised training using both $\mathbf{I}_{\mathrm{A,T}}$ and $\mathbf{I}_{\mathrm{B}}$ (see Fig.~\ref{fig:prob_redef}b). 
This enables the homography estimation network to implicitly learn cross-modal homography knowledge at the beginning of the training, and triggers the two split optimized sub-problems to converge into our desired direction, as shown by the orange curve in Fig.~\ref{fig:opt_MACE}a.


\section{Method}\label{sec:method}
Based on the problem reformulation in Section~\ref{sec:reformulation}, we further investigate its potential and propose an unsupervised cross-modal homography estimation learning framework, named \textbf{S}plit \textbf{S}upervised \textbf{H}omography estimation \textbf{N}etwork (SSHNet). Section~\ref{sec:framework} covers the split optimization of SSHNet (Fig.~\ref{fig:method}a and Fig.~\ref{fig:method}b), Section~\ref{sec:homography_network} discusses the cooperation of SSHNet with various homography estimation architectures, Section~\ref{sec:transfer_network} outlines the design of the modality transfer network, Section~\ref{sec:extra_sup} presents the extra homography feature space supervision (Fig.~\ref{fig:method}c), and Section~\ref{sec:distillation} illustrates the distillation training technique (Fig.~\ref{fig:distillation}).

\subsection{Split Optimization}\label{sec:framework}
As illustrated in Fig.~\ref{fig:method}a and Fig.~\ref{fig:method}b, the homography estimation network and the modality transfer network are optimized separately in each iteration, forming the split optimization framework of our SSHNet. In the following, we will elaborate the optimization procedure for each network.

\textbf{Optimization of Sub-problem \romannum{1}.} As mentioned in Section~\ref{sec:reformulation}, we adopt a two-branch self-supervised learning strategy to optimize the homography estimation network in Sub-problem \romannum{1}. We apply the separate simulated homography deformation on $\mathbf{I}_{\mathrm{A}}$ and $\mathbf{I}_{\mathrm{B}}$, generating the intra-modal image pairs $(\mathbf{I}_{\mathrm{A}}, \mathbf{I}_{\mathrm{A}}^\prime)$ and $(\mathbf{I}_{\mathrm{B}}, \mathbf{I}_{\mathrm{B}}^\prime)$ with corresponding ground-truth homography $\mathbf{H}_{\mathrm{A,GT}}$ and $\mathbf{H}_{\mathrm{B,GT}}$. The modality transfer network transfers $(\mathbf{I}_{\mathrm{A}}, \mathbf{I}_{\mathrm{A}}^\prime)$ into $(\mathbf{I}_{\mathrm{A,T}}, \mathbf{I}_{\mathrm{A,T}}^\prime)$. We simultaneously use $(\mathbf{I}_{\mathrm{B}}, \mathbf{I}_{\mathrm{B}}^\prime)$ and $(\mathbf{I}_{\mathrm{A,T}}, \mathbf{I}_{\mathrm{A,T}}^\prime)$ to train the weight-shared homography estimation network in a self-supervised manner. The optimization of Sub-problem \romannum{1} can be formulated as:
\begin{equation}
    \underset{\zeta}{\arg \min } \ L_\mathrm{H}\big(\mathcal{H}_\zeta(\mathbf{I}_{\mathrm{B}}, \mathbf{I}_{\mathrm{B}}^\prime), \mathbf{H}_{\mathrm{B,GT}} \big) + \mathcal{R}\big(\mathbf{I}_{\mathrm{A,T}}, \mathbf{I}_{\mathrm{A,T}}^\prime \big), \\
\end{equation} 
where $\mathcal{H}_\zeta$ denotes the homography estimation network with parameters $\zeta$ to be optimized, and $L_\mathrm{H}$ denotes the homography estimation loss. $\mathcal{R}(\mathbf{I}_{\mathrm{A,T}}, \mathbf{I}_{\mathrm{A,T}}^\prime) = L_\mathrm{H}\big(\mathcal{H}_\zeta(\mathcal{T}_{\theta^\ast}(\mathbf{I}_{\mathrm{A}}), \mathcal{T}_{\theta^\ast}(\mathbf{I}_{\mathrm{A}}^\prime)), \mathbf{H}_{\mathrm{A,GT}} \big)$, which is the regularization term, where $\mathcal{T}_{\theta^\ast}$ denotes the modality transfer network with parameters $\theta^\ast$ frozen.

\textbf{Optimization of Sub-problem \romannum{2}.} We train a modality transfer network to convert $\mathbf{I}_{\mathrm{A}}$ into an image of pseudo modality B, namely $\mathbf{I}_{\mathrm{A,T}}$. Following the reformulation in Section~\ref{sec:reformulation}, we apply the homography matrix $\hat{\mathbf{H}}_\mathrm{AB}$ predicted by $\mathcal{H}_\zeta$ to warp $\mathbf{I}_{\mathrm{B}}$, providing direct supervision for the modality transfer network. The optimization of Sub-problem \romannum{2} can be formulated as
\begin{equation}
    \underset{\theta}{\arg \min } \ L_\mathrm{T}\big(\mathcal{T}_\theta(\mathbf{I}_{\mathrm{A}}), \mathcal{W}(\mathbf{I}_{\mathrm{B}}, \mathcal{H}_{\zeta^\ast}(\mathbf{I}_{\mathrm{A,T}}, \mathbf{I}_{\mathrm{B}}))\big),
\end{equation}
where $\mathcal{T}_\theta$ denotes the modality transfer network with parameters $\theta$ to be optimized, $\mathcal{H}_{\zeta^\ast}$ denotes the homography estimation network with parameters $\zeta^\ast$ frozen, $\mathcal{W}$ denotes the image warping with the estimated homography, 
and $L_\mathrm{T}$ denotes the modality transfer loss.

\subsection{Homography Estimation Network}\label{sec:homography_network}
Present homography estimation approaches exhibit a variety of architectural designs. Some approaches~\cite{detone2016deep,le2020deep,nguyen2018unsupervised,zhang2020content,ye2021motion} concatenate input images in the channel dimension, directly estimating the homography deformation. Others~\cite{shao2021localtrans, cao2022iterative,cao2023recurrent,zhang2024scpnet} use backbone networks to extract features and employ the feature correlation to perform homography decoding. Additionally, recent approaches~\cite{cao2022iterative, cao2023recurrent} further introduce iterative architectures to refine the estimation results, achieving substantial improvement in accuracy. 

As illustrated in the experimental results of Section~\ref{sec:comparisons}, our proposed framework is independent of the specific homography estimation architecture, enabling integration with various architectures. The pure supervised training strategy significantly boosts the stability of the optimization process. As a result, iterative homography estimation architectures like IHN~\cite{cao2022iterative} and RHWF~\cite{cao2023recurrent} can be effectively employed, leading to a higher performance upper bound. In contrast, due to the lack of an explicit solution for consistent features between cross-modal images, the previous SOTA approach, SCPNet~\cite{zhang2024scpnet}, struggles to converge when using these iterative architectures.
For the self-supervised training, we adopt $L_1$ homography estimation loss following previous works~\cite{detone2016deep,le2020deep,cao2022iterative,cao2023recurrent,zhang2024scpnet}, which is formulated as
\begin{equation}
    L_\mathrm{H}(\hat{\mathbf{H}}, \mathbf{H}_\mathrm{GT}) = \left\|\hat{\mathbf{H}}-\mathbf{H}_{\mathrm{GT}}\right\|_1,
\end{equation}
where $\hat{\mathbf{H}}$ denotes the estimated homography, ${\mathbf{H}}_\mathrm{GT}$ denotes the ground-truth homography. 
For iterative homography architectures, we use a weighted sum of the $L_1$ distance following ~\cite{cao2022iterative,cao2023recurrent}.

\subsection{Modality Transfer Network}\label{sec:transfer_network}
We design the modality transfer network in our SSHNet as a U-shaped architecture to enhance the perception of multi-scale information. Inspired by Swin-Unet~\cite{cao2022swin}, the Swin Transformer block~\cite{liu2021swin} serves as the basic unit of the modality transfer network. The attention mechanism enhances the network's ability to capture hierarchical structures and context information. This facilitates more effective learning of complex mappings between modalities. 

As mentioned in Section~\ref{sec:framework}, we warp image $\mathbf{I}_{\mathrm{B}}$ with the $\hat{\mathbf{H}}_\mathrm{AB}$ predicted by the homography estimation network for the modality transfer supervision. 
During optimization, $\hat{\mathbf{H}}_\mathrm{AB}$ cannot ideally align the input image $\mathbf{I}_\mathrm{A}$ with the supervisory image $\mathbf{I}_\mathrm{B}$.
Therefore, instead of using pixel-wise $L_1$ or $L_2$ loss, we apply perceptual loss $L_\mathrm{pcp}$~\cite{johnson2016perceptual} to capture high-level representations, ensuring better tolerance to misalignments, which is formulated as
\begin{equation}
    L_\mathrm{pcp}(\mathbf{I}_\mathrm{A,T}, \mathbf{I}_{\mathrm{B,W}}) = \sum_j \frac{1}{C_j H_j W_j} \left\|\psi_j(\mathbf{I}_\mathrm{A,T})-\psi_j(\mathbf{I}_{\mathrm{B,W}})\right\|_2^2,
\end{equation}
where $\psi_j$ is the $j$-th layer of the pre-trained VGG with a feature map of shape $C_j \times H_j \times W_j$, $\mathbf{I}_{\mathrm{B,W}}$ denotes the image $\mathbf{I}_{\mathrm{B}}$ warped by the estimated $\hat{\mathbf{H}}_\mathrm{AB}$.

\subsection{Extra Homography Feature Space Supervision}\label{sec:extra_sup}
The current modality transfer network in Sub-problem \romannum{2} is supervised by the perceptual loss~\cite{johnson2016perceptual}, which computes feature similarity using a pre-trained VGG network.
This may yield suboptimal results for homography estimation in Sub-problem \romannum{1} when trained under the present totally split optimization framework. 

To address this issue, as illustrated in Fig.~\ref{fig:method}c, we introduce an extra supervision in the homography feature space when the architecture of homography estimation network has explicit feature extractors, such as IHN, RHWF, and SCPNet. A homography feature (HF) loss that directly enhances the feature consistency for homography estimation between $\mathbf{I}_{\mathrm{A,T}}$ and $\mathbf{I}_{\mathrm{B}}$ is computed. The modality transfer network and the feature extractor of the homography estimation network are optimized together under the supervision of HF loss, which can be expressed as
\begin{equation}
    \underset{\theta, \xi}{\arg \min } \ L_\mathrm{HF}\big(\mathcal{F}_\xi(\mathcal{T}_\theta(\mathbf{I}_{\mathrm{A}})), \mathcal{F}_\xi(\mathcal{W}(\mathbf{I}_{\mathrm{B}}, \hat{\mathbf{H}}_\mathrm{AB}))\big),
\end{equation}
where $\mathcal{F}_\xi$ denotes the feature extractor with parameters $\xi$ to be optimized. 
We note that IHN, RHWF, and SCPNet all depend on feature correlation to perform homography decoding. Considering this, We formulate the HF loss using a correlation-based loss function $L_\mathrm{corr}$ to enhance feature consistency, which is formulated as
\begin{equation}
    L_\mathrm{corr}(\mathbf{F}_{\mathrm{A,T}}, \mathbf{F}_{\mathrm{B,W}}) = - \sum_{i, j} {\mathbf{F}_{\mathrm{A,T}}(i,j)}^{\top} \mathbf{F}_{\mathrm{B,W}}(i,j),
\end{equation}
where $\mathbf{F}_{\mathrm{A,T}}$ and $\mathbf{F}_{\mathrm{B,W}}$ denote the feature maps of $\mathbf{I}_\mathrm{A,T}$ and $\mathbf{I}_\mathrm{B,W}$ in the homography estimation network. 

\begin{figure}[t]
    \centering
    \includegraphics[width=0.47\textwidth]{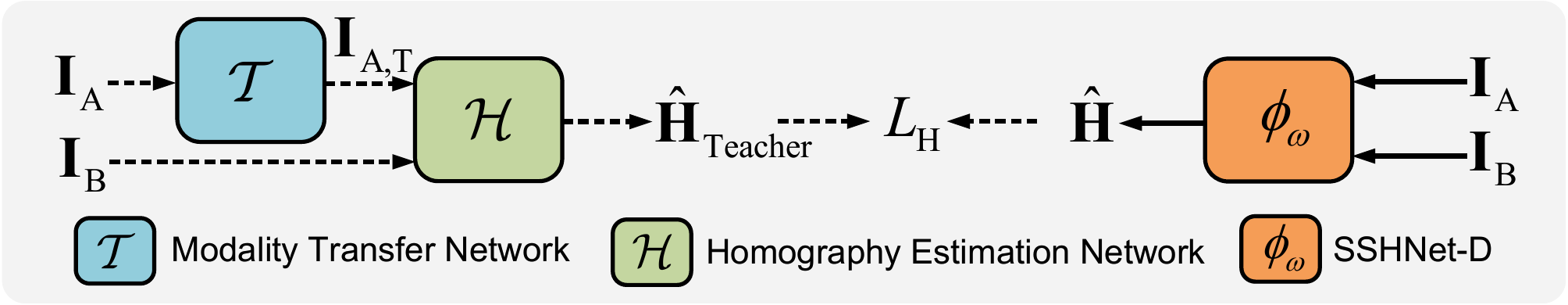}
    \vspace{-0.7em}
    \caption{Schematic diagram of the distillation training.}\label{fig:distillation}
    \vspace{-1.5em}
\end{figure}

\subsection{Distillation Training}\label{sec:distillation}
We can see that the present framework has extra network parameters due to the utilization of the modality transfer network in Sub-problem \romannum{2}. Moreover, as the modality transfer network is trained on multi-modal image pairs from a fixed domain, its generalization ability to images from other domains may be limited.

To cope with the above problem, we introduce a distillation training technique under the inspiration of the unsupervised knowledge distillation frameworks in domain adaptive object detection~\cite{saito2017asymmetric,zhou2023ssda}.
We use SSHNet as a teacher network to guide a student network that only retains the homography estimation network, named Distilled SSHNet (SSHNet-D). During the distillation training, the homography $\hat{\mathbf{H}}_\mathrm{teacher}$ predicted by SSHNet serves as a pseudo ground-truth to supervise SSHNet-D, as illustrated in Fig.~\ref{fig:distillation}. The distillation training can be formulated as
\begin{equation}
    \underset{\omega}{\arg \min } \ L_\mathrm{H}\big(\phi_\omega(\mathbf{I}_{\mathrm{A}}, \mathbf{I}_{\mathrm{B}}), \hat{\mathbf{H}}_\mathrm{teacher} \big),
\end{equation}
where $\phi_\omega$ denotes the SSHNet-D with the parameters $\omega$ to be optimized.

\section{Experiments}\label{sec:experiments}
\subsection{Experimental Settings}
\begin{figure*}[t]
	\centering
	\includegraphics[width=0.98\textwidth]{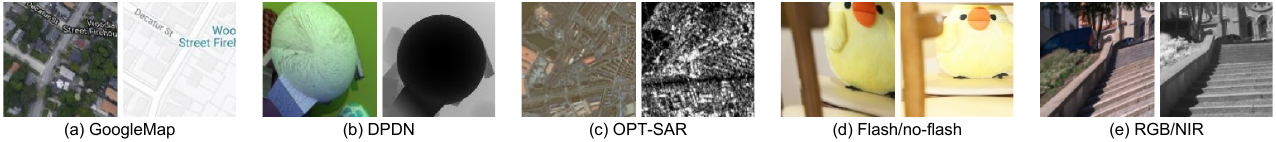}
	\vspace{-0.7em}
	\caption{
		Examples for each dataset.
	}\label{fig:datasets}
	\vspace{-0.7em}
\end{figure*}

\textbf{Datasets.} 
We evaluate SSHNet on various datasets, including GoogleMap~\cite{zhao2021deep}, DPDN~\cite{riegler2016deep}, OPT-SAR~\cite{li2022mcanet}, Flash/no-flash~\cite{he2014saliency} cross-modal datasets, and RGB/NIR~\cite{brown2011multi} cross-spectral dataset. 
Examples for each dataset are shown in Fig.~\ref{fig:datasets}. 
To ensure a fair comparison, all evaluated methods are trained and tested using the same training and test splits within each dataset. 

\textbf{Baselines.} 
We evaluate SSHNet with traditional feature-based approaches including SIFT~\cite{lowe2004distinctive}, ORB~\cite{rublee2011orb}, unsupervised approaches including UDHN~\cite{nguyen2018unsupervised}, biHomE~\cite{koguciuk2021perceptual}, CA-UDHN~\cite{zhang2020content}, BasesHomo~\cite{ye2021motion}, UMF-CMGR~\cite{wang2022unsupervised}, SCPNet~\cite{zhang2024scpnet}, and supervised approaches including DHN~\cite{detone2016deep}, MHN~\cite{le2020deep}, LocalTrans~\cite{shao2021localtrans}, IHN~\cite{cao2022iterative}, RHWF~\cite{cao2023recurrent}. For feature-based approaches, we choose RANSAC~\cite{fischler1981random} as the outlier rejection algorithms. UMF-CMGR is an image fusion approach based on registration, and we use the registration part for comparison. 

\textbf{Homography Estimation Architectures.} 
To investigate the potential of our proposed SSHNet, we adopt various homography estimation architectures including DHN~\cite{detone2016deep} and MHN~\cite{le2020deep}, together with iterative architectures such as IHN~\cite{cao2022iterative} and RHWF~\cite{cao2023recurrent}. Moreover, we also adopt the architecture from the previous SOTA unsupervised approach SCPNet~\cite{zhang2024scpnet}, to specifically evaluate the effectiveness of our proposed unsupervised learning framework.

\textbf{Metrics.}
Similar to previous works~\cite{detone2016deep, zhao2021deep, cao2022iterative, cao2023recurrent, shao2021localtrans}, we randomly perturb the corner points of 128 $\times$ 128 images to make the deformed image, with a perturbation range set to [-32, 32].
We evaluate homography estimation accuracy using the mean average corner error (MACE) following~\cite{detone2016deep, zhao2021deep, cao2022iterative,cao2023recurrent, shao2021localtrans}.
A lower MACE indicates a higher accuracy.

\subsection{Ablation}\label{sec:ablation}
The ablation studies are conducted on the OPT-SAR dataset.
We adopt IHN as the homography estimation network in our SSHNet for ablation studies.

\begin{table}[t]
	\centering
	\caption{
		Ablation study on the training strategy of SSHNet.  
	}\label{Ablation training strategy}
	\vspace{-0.7em}
	\renewcommand\arraystretch{0.95}
	\renewcommand\tabcolsep{27pt}
	\resizebox{\linewidth}{!}
	{
		\begin{tabular}{lc}
			\toprule
			Experiment                      & MACE $\downarrow$ \\
			\midrule
			w/o problem reformulation       & NC \\
			w/o split optimization          & NC \\
			pure self-supervised learning 	& 24.60 \\
			SSHNet                          & \textbf{2.94} \\
			\bottomrule
		\end{tabular}
	}
	\vspace{-0.5em}
\end{table}

\begin{table}[t]
	\centering
	\caption{
		Ablation study on the architectures of the modality transfer network.
	}\label{Ablation architectures}
	\vspace{-0.7em}
	\renewcommand\arraystretch{0.9}
	\renewcommand\tabcolsep{16pt}
	\resizebox{\linewidth}{!}
	{
		\begin{tabular}{lcc}
			\toprule
			Experiment                        & Parameters (M) & MACE $\downarrow$ \\
			\midrule
			CNN-based            & 7.48 &  8.72  \\
			Transformer-based    & 7.54 & \textbf{2.94} \\
			\bottomrule
		\end{tabular}
	}
	\vspace{-0.5em}
\end{table}

\begin{table}[t]
	\centering
	\caption{
		Ablation study on the loss functions. $L_\mathrm{T}$ denotes the modality transfer loss, $L_\mathrm{HF}$ denotes the homography feature loss.
	}\label{Ablation losses}
	\vspace{-0.7em}
	\renewcommand\arraystretch{0.9}
	\renewcommand\tabcolsep{22pt}
	\resizebox{\linewidth}{!}
	{
		\begin{tabular}{llc}
			\toprule
			Experiment & Loss Type    & MACE $\downarrow$ \\
			\midrule
			\multirow{3}{*}{$L_\mathrm{T}$} &   $L_1$       & 5.88 \\
			&   $L_2$       & 6.23 \\
			&   $L_\mathrm{pcp}$   & \textbf{4.52} \\
			\midrule
			\multirow{3}{*}{$L_\mathrm{T} + L_\mathrm{HF}$} &   $L_\mathrm{pcp} + L_1$       & 4.09    \\
			&   $L_\mathrm{pcp} + L_2$       & 3.27 \\
			&   $L_\mathrm{pcp} + L_\mathrm{corr}$  & \textbf{2.94} \\
			\bottomrule
		\end{tabular}
	}
	\vspace{-0.5em}
\end{table}

\begin{table}[t]
	\caption{
		Ablation study of the modality transfer direction.
	}\label{table direction}
	\vspace{-0.7em}
	\centering
	\renewcommand\arraystretch{1.1}
	\renewcommand\tabcolsep{1pt}
	\small
	\resizebox{\linewidth}{!}
	{
		\begin{tabular}{c|cc|cc}
			\toprule
			{\diagbox{Method}{Direction}} & RGB$\rightarrow$NIR & NIR$\rightarrow$RGB & OPT$\rightarrow$SAR & SAR$\rightarrow$OPT \\
			\midrule
			SSHNet & 1.66 & 1.75 & 2.94 & 4.28 \\
			\bottomrule
		\end{tabular}
	}
	\vspace{-1.5em}
\end{table}

\begin{table*}[t]
	\caption{
		Quantitative comparison of our SSHNet and other approaches. 
		NC denotes the training is not converged. 
		The best results of unsupervised methods are \textbf{highlighted in bold}.
	}\label{table evaluation}
	\centering
	\renewcommand\tabcolsep{15pt}
	\renewcommand\arraystretch{0.94}
	\vspace{-0.5em}
	\resizebox{\linewidth}{!}
	{
		\begin{tabular}{c|l|ccccc}
			\toprule
			\multicolumn{2}{c|}{\diagbox{Approach}{Dataset}} & GoogleMap & DPDN & OPT-SAR & Flash/no-flash & RGB/NIR \\
			\midrule
			\multirow{2}{*}{Traditional}    & SIFT~\cite{lowe2004distinctive}      & 24.53 & 25.14 & 24.82 & 18.29 & 21.33 \\
			& ORB~\cite{rublee2011orb}             & 24.52 & 24.86 & 24.90 & 20.22 & 22.67 \\
			\midrule
			\multirow{5}{*}{Supervised}     & DHN~\cite{detone2016deep}            & 5.20  & 4.92  & 8.27  & 6.42  & 11.88 \\
			& MHN~\cite{le2020deep}                & 2.69  & 2.95  & 5.59  & 5.24  & 5.52  \\
			& LocalTrans~\cite{shao2021localtrans} & 3.22  & 1.79  & 20.68 & 1.52  & 1.72  \\
			& IHN~\cite{cao2022iterative}          & 0.92  & 1.17  & 1.67  & 0.80  & 1.63  \\
			& RHWF~\cite{cao2023recurrent}         & 0.71  & 1.06  & 1.48  & 0.65  & 1.07  \\
			\midrule
			\multirow{12}{*}{Unsupervised}  & UDHN~\cite{nguyen2018unsupervised}   & 22.84 & NC    & NC    & 21.20 & 23.43 \\
			& CA-UDHN~\cite{zhang2020content}      & 24.61 & 24.99 & 24.76 & 21.32 & 24.12 \\                                
			& biHomE~\cite{koguciuk2021perceptual} & NC    & NC    & NC    & 11.86 & 23.77 \\
			& BasesHomo~\cite{ye2021motion}        & 24.49 & 27.33 & 26.72 & 25.12 & 24.41 \\
			& UMF-CMGR~\cite{wang2022unsupervised} & 24.60 & 25.46 & 24.70 & 23.49 & 22.38 \\
			& SCPNet~\cite{zhang2024scpnet}        & 4.35  & NC    & NC    & 2.67  & 4.78  \\
			
			& \cellcolor{gray!20}SSHNet-DHN (Ours)  & \cellcolor{gray!20}9.28  & \cellcolor{gray!20}9.71  & \cellcolor{gray!20}17.63 & \cellcolor{gray!20}10.51 & \cellcolor{gray!20}12.13 \\
			& \cellcolor{gray!20}SSHNet-MHN (Ours)                  & \cellcolor{gray!20}2.90  & \cellcolor{gray!20}3.04  & \cellcolor{gray!20}6.77  & \cellcolor{gray!20}5.62  & \cellcolor{gray!20}6.93  \\
			& \cellcolor{gray!20}SSHNet-SCPNet (Ours)               & \cellcolor{gray!20}3.89  & \cellcolor{gray!20}4.81  & \cellcolor{gray!20}12.94 & \cellcolor{gray!20}2.65  & \cellcolor{gray!20}3.86  \\
			& \cellcolor{gray!20}SSHNet-RHWF (Ours)                 & \cellcolor{gray!20}1.29  & \cellcolor{gray!20}1.26  & \cellcolor{gray!20}3.08  & \cellcolor{gray!20}\textbf{0.95} & \cellcolor{gray!20}\textbf{1.52} \\
			& \cellcolor{gray!20}SSHNet-IHN (Ours)                  & \cellcolor{gray!20}\textbf{1.23} & \cellcolor{gray!20}\textbf{1.12} & \cellcolor{gray!20}\textbf{2.94}  & \cellcolor{gray!20}1.08 & \cellcolor{gray!20}1.66 \\
			& \cellcolor{gray!40}SSHNet-IHN-D (Ours)                & \cellcolor{gray!40}1.26  & \cellcolor{gray!40}1.24  & \cellcolor{gray!40}3.31 & \cellcolor{gray!40}1.16  & \cellcolor{gray!40}1.72  \\
			\bottomrule
		\end{tabular}
	}
	\vspace{-1.2em}
\end{table*}

\textbf{Training Strategy.} 
We conduct an ablation study on the training strategy of our SSHNet.
The experimental results in Table~\ref{Ablation training strategy} demonstrate that both problem reformulation and the split optimization strategy are crucial for SSHNet. Without problem reformulation, unsupervised training with intensity-based supervision fails to converge. Even with problem reformulation, directly optimizing both networks simultaneously does not yield reliable homography estimation. SSHNet achieves effective convergence only when problem reformulation is combined with the split optimization strategy, where each network is trained separately.
We also compare our framework with method trained through pure self-supervised learning. Results in Table~\ref{Ablation training strategy} show that self-supervised learning alone does not yield satisfactory homography estimation results.

\textbf{Architectures of the Modality Transfer Network.} As illustrated in Table~\ref{Ablation architectures}. Our Transformer-based architecture outperforms the CNN-based architecture by a large margin, with similar parameter counts.

\textbf{Loss Functions.}
To further investigate the effectiveness of the modality transfer and homography feature loss functions, we first train the network without HF loss to evaluate the performance of different types of modality transfer loss.
The results in the upper part of Table~\ref{Ablation losses} indicate that the perceptual loss $L_\mathrm{pcp}$ outperforms $L_1$ and $L_2$.
Then we adopt $L_\mathrm{pcp}$ as the modality transfer loss and add different types of HF loss to analyze their impact.
We report the results in the lower part of Table~\ref{Ablation losses}. 
Though all different types of HF loss further boost the network's performance, the correlation-based loss $L_\mathrm{corr}$ achieves superior performance compared to $L_1$ and $L_2$. 

\textbf{Modality transfer direction.} 
Table~\ref{table direction} presents the results of different transfer directions on the RGB/NIR and OPT-SAR datasets. Since some modalities carry more information, transferring from richer to less informative modalities facilitates the modality transfer network's learning process, thereby promoting the framework's convergence.

\subsection{Comparisons with Existing Methods}\label{sec:comparisons}
We compare the homography estimation results of our proposed SSHNet and baselines and list the experimental results in Table~\ref{table evaluation}. 
It is observed that the prior traditional and unsupervised approaches can hardly handle the homography estimation under large modality and deformation gaps. 
In contrast, our method achieves much more satisfactory results with various homography estimation architectures. 
When adopting the iterative architectures IHN and RHWF, our method outperforms all the compared traditional and unsupervised approaches by a significant margin. 
Moreover, even though trained without direct cross-modal supervision, SSHNet-IHN and SSHNet-RHWF outperforms supervised approaches such as DHN, MHN and LocalTrans. Specifically, SSHNet-IHN owns 54.3\% and 47.4\% lower MACEs than MHN, and 61.8\% and 85.8\% lower MACEs than LocalTrans on the challenging GoogleMap and OPT-SAR datasets, respectively. 
Notably, SSHNet-SCPNet, which uses the homography estimation architecture from the previous unsupervised SOTA approach SCPNet, outperforms SCPNet itself on all datasets, further proving the superiority of our proposed training framework.

We also conduct the distillation training on SSHNet with different homography estimation architectures. We present the distillation training results of SSHNet-IHN in Table~\ref{table evaluation}, namely SSHNet-IHN-D. We observe that the performance of SSHNet-IHN-D is comparable to that of SSHNet-IHN. 
For more detailed qualitative visualization and experimental results, please refer to supplementary material.

\subsection{Cross-Dataset Generalization}\label{sec:cross-dataset}
We further conduct a cross-dataset evaluation of SSHNet-IHN-D and SSHNet-IHN on the aforementioned datasets, to further reveal the effectiveness of our proposed distillation training. As illustrated in Fig.~\ref{fig:datasets}, multi-modal image pairs from the GoogleMap, DPDN, and OPT-SAR datasets exhibit substantial modality gaps, and we call these three datasets challenging datasets, for a better illustration.
In contrast, the Flash/no-flash and RGB/NIR datasets show smaller modality gaps, so we call them easy datasets. For each dataset, we compute the percentage of homography predictions that have average corner errors (ACE) lower than 5 pixels on the test set, which reveals the ratio of reliable estimations. We directly list the differences of the percentage of SSHNet-IHN-D and SSHNet-IHN in Table~\ref{table cross dataset}.

\begin{table}[t]
    \caption{
        Cross-dataset evaluation results. 
        \textcolor{red}{Red} indicates the improved performance of SSHNet-IHN-D compared to SSHNet-IHN, and \textcolor{blue}{blue} indicates the declined 
        one.
    }\label{table cross dataset}
    \vspace{-0.7em}
    \centering
    \renewcommand\arraystretch{1.2}
    \renewcommand\tabcolsep{5pt}
    \small
    \resizebox{\linewidth}{!}
    {
        \begin{tabular}{c|ccccc}
            \toprule
            {\diagbox{Train}{Test}} & GoogleMap & DPDN & OPT-SAR & Flash/no-flash & RGB/NIR\\
            \midrule
            GoogleMap    & /  & \textcolor{red}{$\uparrow 1.8\%$}  & \textcolor{red}{$\uparrow 1.0\%$}  & \textcolor{red}{$\uparrow 80.4\%$} & \textcolor{red}{$\uparrow 63.0\%$}  \\
            DPDN     & \textcolor{red}{$\uparrow 66.5\%$} & / & \textcolor{red}{$\uparrow 0.4\%$}  & \textcolor{red}{$\uparrow 92.1\%$} & \textcolor{red}{$\uparrow 75.8\%$} \\
            OPT-SAR      & \textcolor{red}{$\uparrow 48.4\%$} & \textcolor{red}{$\uparrow 1.3\%$}  & / & \textcolor{red}{$\uparrow 77.5\%$}  & \textcolor{red}{$\uparrow 66.2\%$}\\
            Flash/no-flash      & \textcolor{blue}{$\downarrow 5.2\%$}  & \textcolor{blue}{$\downarrow 0.2\%$} & \textcolor{red}{$\uparrow 0.1\%$}  & / & \textcolor{blue}{$\downarrow 4.6\%$} \\
            RGB/NIR     & \textcolor{blue}{$\downarrow 0.5\%$} & \textcolor{red}{$\uparrow 1.2\%$} & \textcolor{red}{$\uparrow 3.0\%$} & \textcolor{red}{$\uparrow 2.1\%$} & / \\
            \bottomrule
        \end{tabular}
    }
    \vspace{-1.5em}
\end{table}

SSHNet-IHN uses a modality transfer network to transfer images from one modality to the other on the training dataset, making it highly prone to overfit the training data. 
In contrast, SSHNet-IHN-D excludes the modality transfer network and directly extracts consistent features from cross-modal images, and hence is forced to learn the features that is robust across different domains. As illustrated in Table~\ref{table cross dataset}, when trained on challenging datasets, SSHNet-IHN-D exhibits significant superior generalization ability. Specifically, SSHNet-IHN-D trained on DPDN yields 66.5\%, 92.1\%, and 75.8\% improvement on GoogleMap, Flash/no-flash and RGB/NIR datasets compared to SSHNet. We note that the generalization ability across the challenging datasets remains limited due to the substantial domain gaps between the datasets. As for training on easy datasets, the generalization performance improvement is plain. This is because the training data is too easy to provide enough knowledge for obtaining the features robust across different domains.

\subsection{Real World Scenarios} \label{sec:real_world}
\begin{figure}[t]
	\centering
	\includegraphics[width=0.48\textwidth]{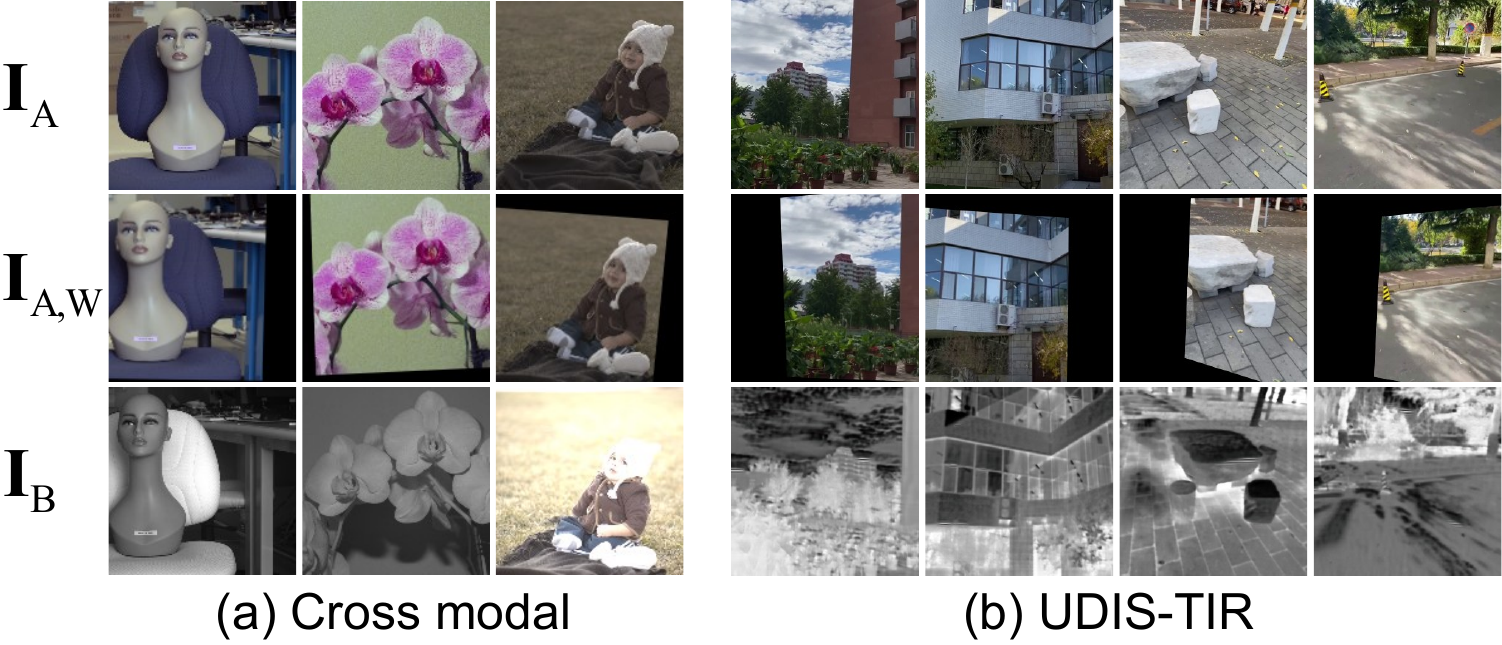}
	\vspace{-1.8em}
	\caption{Visualization of results on real-world deformations.}
	\label{fig:vis}
	\vspace{-0.7em}
\end{figure}

We further evaluate our method on real-world scenarios~\cite{shen2014multi} that do not satisfy perfect homography alignment, including RGB/NIR and Flash/no-flash image pairs. As illustrated in Fig.~\ref{fig:vis}a, the model trained with simulated homographies achieves effective performance. 
We also use a pretrained RGB-to-TIR translation network to convert RGB-RGB image pairs from the UDIS~\cite{nie2021udis} dataset into RGB-TIR image pairs, creating the UDIS-TIR dataset with parallax image pairs. Following UDIS, we evaluate SSHNet-IHN on the UDIS-TIR dataset and compute the PSNR and SSIM on RGB images before modality transfer.
We present the quantitative results in Table~\ref{table real} and show more visualization results in Fig.~\ref{fig:vis}b.

\begin{table}[t]
	\caption{Experimental results on the UDIS-TIR dataset. $I_{3\times3}$ denotes the identity matrix.}\label{table real}
	\vspace{-0.7em}
	\centering
	\renewcommand\arraystretch{0.95}
	\renewcommand\tabcolsep{25pt}
	\small
	\resizebox{\linewidth}{!}{
		\begin{tabular}{c|cc}
			\toprule
			Method & PSNR $\uparrow$ & SSIM $\uparrow$  \\
			\midrule
			$I_{3\times3}$ & 12.92 & 0.306 \\
			SCPNet & NC  & NC  \\
			SSHNet-IHN & 20.05 & 0.728 \\
			\bottomrule
		\end{tabular}
	}
	\vspace{-1.5em}
\end{table}

\section{Conclusions}
In this paper, we propose SSHNet, a novel unsupervised cross-modal homography estimation framework. SSHNet reformulates the problem into two supervised sub-problems, addressed by a homography estimation network and a modality transfer network. These networks are trained separately using a split optimization strategy, ensuring stable training.
We also introduce an extra homography feature space supervision to further improve feature consistency during the homography estimation and boost the performance of our framework.
Moreover, we employ a distillation training technique to reduce model parameters and improve cross-domain generalization ability.
Experimental results show that our SSHNet achieves SOTA performance on multiple datasets among unsupervised approaches, and outperforms many supervised approaches.

\noindent\textbf{Acknowledgments.}
This work was supported in part by the Zhejiang Provincial Natural Science Foundation of China under grant LD24F020003, in part by the National Natural Science Foundation of China under grant 62301484, in part by the Ningbo Natural Science Foundation of China under grant 2024J454, and in part by the National Key Research and Development Program of China under grant 2023YFB3209800. We also thank the generous help from Jianxin Hu, Zhejiang University.
{
    \small
    \bibliographystyle{ieeenat_fullname}
    \bibliography{main}
}


\end{document}